%% file: TMM_main.tex
\documentclass[journal]{IEEEtran}
\input{macro}

\ifCLASSINFOpdf
\else
\fi
\begin{document}
%
\title{Ensemble Learning with Manifold-Based Data Splitting for Noisy Label Correction}
%
%
%

\author{Hao-Chiang~Shao,~\IEEEmembership{Member,~IEEE,}
        Hsin-Chieh~Wang,
        Weng-Tai~Su,
        and~Chia-Wen~Lin,~\IEEEmembership{Fellow,~IEEE}
\thanks{Manuscript received February 28, 2021; revised ** **, ****. This work was supported in part by Ministry of Science and Technology, Taiwan, under Grant 110-2634-F-007-015-. (Corresponding author:
Chia-Wen Lin)}
\thanks{H.-C. Shao is with the Department of Statistics and Information Science, Fu Jen Catholic University, New Taipei City, Taiwan. (email: shao.haochiang@gmail.com)}
\thanks{H.-C.~Wang and W.-T.~Su are with the Department of Electrical Engineering, Nation Tsing Hua University, Hsinchu, Taiwan. (email: lucas85062055@gmail.com; s105061805@m105.nthu.edu.tw )}
\thanks{C.-W.~Lin is with the Department of Electrical Engineering and the Institute of Communication Engineering, Nation Tsing Hua University, Hsinchu, Taiwan. (email: cwlin@ee.nthu.edu.tw)}
}

%
%

\markboth{IEEE Transactions on Multimedia (Manuscript for Review)}%
{Shell \MakeLowercase{\textit{et al.}}: Bare Demo of IEEEtran.cls for IEEE Journals}
%



\maketitle

\begin{abstract}
\input{TMM_sec00_abstract}
\end{abstract}


%
\IEEEpeerreviewmaketitle

	\section{Introduction}
	\label{sec:intro}
	\input{TMM_sec01_intro}

	\section{Related Work}
	\label{sec:related}
	\input{TMM_sec02_review}

	\section{Proposed Method}
	\label{sec:method}
	\input{TMM_sec03_method}

	\section{Experimental Results}
	\label{sec:experiments}
	\input{TMM_sec04_experiment}

	\section{Conclusion}
	\label{sec:conclusion}
	\input{TMM_sec05_conclusion}

\ifCLASSOPTIONcaptionsoff
  \newpage
\fi



%

\bibliographystyle{IEEEtran}
\bibliography{ref}

%

\end{document}

%% file: macro.tex
\usepackage{color}
\usepackage{bbding}
\usepackage{pifont}
\usepackage{wasysym}
\usepackage{amssymb}
\usepackage{amsthm}
\usepackage{booktabs}
\usepackage{multirow} 
\usepackage{setspace}
\usepackage{epsfig}
\usepackage{graphicx}
\usepackage{subfigure}
\usepackage{algorithm}
\usepackage{algpseudocode}
\usepackage{url}
\usepackage{changes}

\usepackage{array}
\newcolumntype{L}[1]{>{\raggedright\let\newline\\\arraybackslash\hspace{0pt}}m{#1}}
\newcolumntype{C}[1]{>{\centering\let\newline\\\arraybackslash\hspace{0pt}}m{#1}}
\newcolumntype{R}[1]{>{\raggedleft\let\newline\\\arraybackslash\hspace{0pt}}m{#1}}

\usepackage{cite}
\usepackage{url}
\usepackage[hidelinks]{hyperref}
\hypersetup{
    colorlinks=true,%
    urlcolor=magenta
}

%% file: TMM_sec00_abstract.tex
Label noise in training data can significantly degrade a model's generalization performance for supervised learning tasks. Here we focus on the problem that noisy labels are primarily mislabeled samples, which tend to be concentrated near decision boundaries, rather than uniformly distributed, and whose features should be equivocal. To address the problem, we propose an ensemble learning method to correct noisy labels by exploiting the local structures of feature manifolds. Different from typical ensemble strategies that increase the prediction diversity among sub-models via certain loss terms, our method trains sub-models on disjoint subsets, each being a union of the nearest-neighbors of randomly selected seed samples on the data manifold.  As a result, each sub-model can learn a coarse representation of the data manifold along with a corresponding graph. Moreover, only a limited number of sub-models will be affected by locally-concentrated noisy labels. The constructed graphs are used to suggest a series of label correction candidates, and accordingly, our method derives label correction results by voting down inconsistent suggestions. Our experiments on real-world noisy label datasets demonstrate the superiority of the proposed method over existing state-of-the-arts.

%% file: TMM_sec01_intro.tex
Learning from noisy data is a vital issue in representation learning for two reasons generally. 
First of all, 
as revealed in \cite{settles2009active}, for a classification model $\Phi(\cdot, \theta)$ learned by optimizing its parameter $\theta$ via the cross-entropy loss on the predicted label, its expected error is affected by i) the label noise, ii) the selected learning model $\Phi$, and iii) the model parameter $\theta$ learned in the training process. Then, when the learning model $\Phi$---suitable or not---is determined, the influence caused by both $\Phi$ and $\theta$ is fixed. Hence, the only factor that may downgrade the model performance becomes the label noise of the ground-truth label $y_i$ of each sample $x_i$.
Second, because it is expensive and time-consuming to collect data with reliably clean annotations, in real-world applications people usually need to exploit low-cost open datasets, e.g., those collected by search engines and labeled by quick annotators or crowd-sourcing, for training purposes. These low-cost datasets usually contain a certain amount of noisy labels and thus are not as reliable as professionally-labelled datasets like COCO \cite{lin2014microsoft} and ImageNet \cite{deng2009imagenet}. Consequently, 
how to learn from noisy data, including how to correct noisy data, has become a crucial task in real-world applications. 

\begin{figure}
    \centering
    \includegraphics[width=0.48\textwidth]{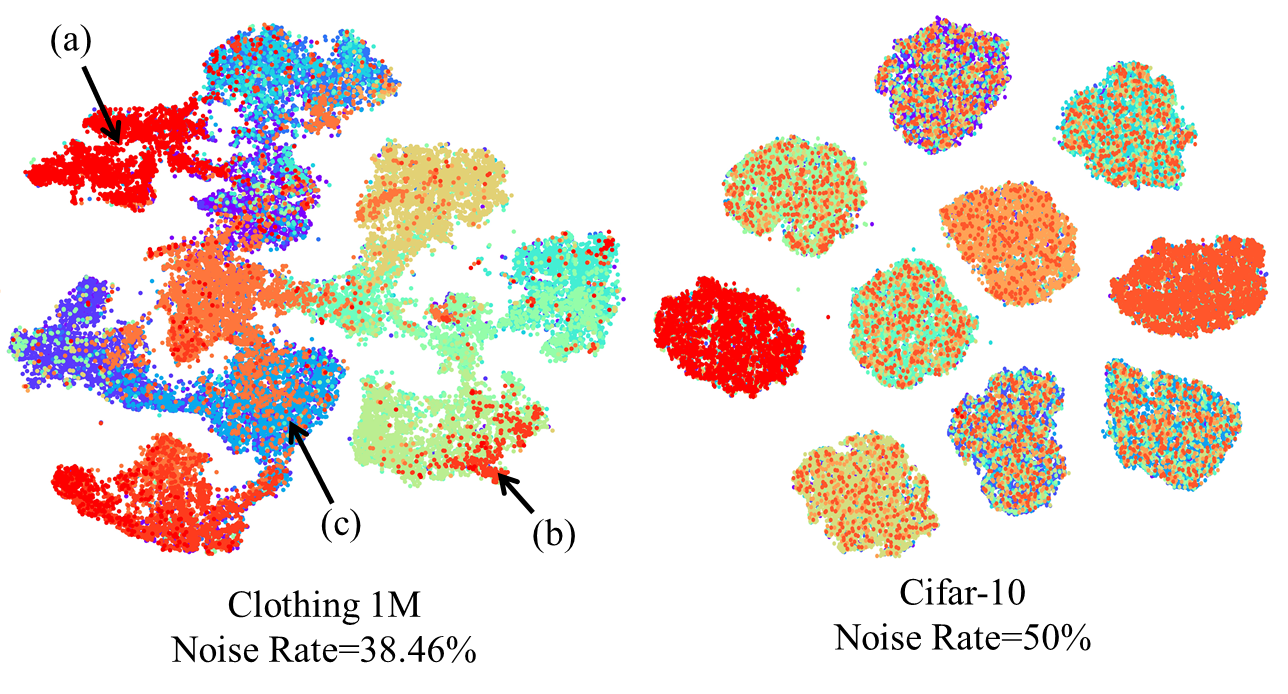}
    \caption{
    Illustration of Confusing label noise versus random label noise. Left: t-SNE map of instance features of Clothing1M. Right: t-SNE map of random perturbed CIFAR-10. In real-world applications, noisy labels are locally concentrated near decision borders, as indicated by arrows (b) and (c), and random noises shown in cIFAR-10 simulations are rare in real cases, as indicated by arrow (a).  
    }
    \label{fig:myfig01}
    \vspace{-0.15 in}
\end{figure}

Label noise can be roughly divided into two types: random label noise and confusing label noise, as illustrated in Fig, \ref{fig:myfig01}. 
The former typically involves mismatched descriptions or tags  usually due to the negligence of an annotator. 
For this type of error, not only a label error may occur randomly in the sample space, but also the erroneous label is often of another irrelevant random class. In contrast, the latter usually occurs when a to-be-labeled sample contains confusing content or equivocal features and is the main cause of noisy labels in real-world applications. Confusing label noise often occurs on data samples lying near the decision boundaries, and such noisy labels should be corrected as one neighboring category to the current one in the feature space.

%
Most existing methods focused on correcting random label noises and reported promising performance on simulated datasets like noisy MNIST \cite{lecun1998gradient} and noisy CIFAR-10 \cite{krizhevsky2009learning} obtained by randomly altering data labels.
%
%
%
%
Primarily, there are two sorts of strategies for solving the noisy label correction problem: transition matrix-based approaches \cite{patrini2017making, hendrycks2018using, xia2019anchor} and class-prototype-based ones \cite{sharma2020noiserank, han2019deep}. 
All these methods learn from a given noisy dataset and correct noisy labels according to the distribution of labels of instances within a local neighborhood. 
As a result, they cannot well cope with real-world confusing label noise robustly because a confusing label is expected to lie near the decision boundary between categories and may be surrounded by other instances with incorrect confusing labels. Moreover, these methods may also give a wrong suggestion to a clean instance once it is located in the transition area between two akin classes in the feature space.

In contrast, some other methods resort to noise-robust loss functions to fight against label noise \cite{tanaka2018joint, zhang2018generalized, wei2020combating,chen2017sentiment}. 
Such noise-robust loss functions operate based on the assumption that whether an instance is mislabeled is irrelevant to its content.
%
That is to say, the distribution of noisy labels is still assumed to be independent of that of data instances. Because this assumption is only valid in the case of random label noise, these methods usually cannot perform well on real-world data involving confusing label noise.


Still, a few label correction methods are ensemble learning-based  approaches \cite{li2019learning, li2020dividemix, nguyen2019self}. By taking a majority decision on predictions of multiple sub-models, ensemble learning approaches can improve the robustness of predictions. 
Some ensemble learning approaches are based on semi-supervised learning strategies, 
aiming to increase the freedom of predictions via suitable data augmentation schemes  \cite{laine2016temporal,tarvainen2017mean}. 
Nevertheless, common data augmentation schemes cannot synthesize data distributing far more differently from the original dataset, especially for those with confusing noise labels. Therefore, these approaches usually need to collaborate with additional loss functions to guarantee the correctness of soft-labels predicted by different sub-networks.  
%
%
Although these methods were designed to increase the uncertainty of sub-networks' predicted labels on noisy samples, they may also decrease the confidence of sub-networks' predictions on clean instances. Hence, corrections suggested by these methods are usually considered as soft-labels for training another student model. A primary drawback of these methods is that they did not take the within-neighborhood label consistency into account, and thus the obtained soft-labels may still be noisy. 
%
%

To address the above problems with confusing label noises, we propose an ensemble-based label correction algorithm by exploiting the local structures of data manifolds. 
As illustrated in Fig. \ref{fig:fig01}, our noisy label correction scheme involves three iterative phases: i) $k$-NN based data splitting, ii) multi-graph label propagation, and iii) confidence-guided label correction. . 
Because confusing label noises tend to be located densely near decision boundaries, they may destroy the label smoothness  within some ``inter-class transient bands'' on the graph locally. By partitioning the source noisy dataset into disjoint subsets using our $k$-NN splitting scheme, each noisy label, along with its $k$-nearest-neighbors, will usually affect only a minority of the ensemble branches. As a result, each ensemble branch generates a graph that holds its own noisy local manifold structures so that such singulars can be treated as outliers during the majority decision process. 
 To this end, we train the  ensemble branches on the corresponding disjoint subsets independently. Through this design, each sub-network can learn not only a coarse global representation of the data manifold, but also different local manifold structures. We then derive label correction suggestions for each sample based on the predictions of the sample's nearest-neighbors in individual disjoint subsets via the corresponding ensemble branches.
Finally, our method suggests final label corrections by ruling out inconsistent suggestions derived according to graphs accessed by ensemble branches. Extensive experiments show the superiority of our method over existing state-of-the-arts.

The novelty of the proposed method is threefold. \\
$\bullet$ We propose a novel iterative data splitting method to split training samples into disjoint subsets, 
each preserving some local manifold structure of source data while representing a coarse global approximation. This design allows the influence of mislabeled instances to be limited to a minority of ensemble branches. \\
$\bullet$ Our design contains a novel noisy-label branch that can stably provide a correct suggestion for within-class clean labels. Hence, this branch can boost the accuracy of label correction result, especially for datasets primarily containing  confusing label noises. \\
$\bullet$ We adopt multi-graph label propagation, rather than a simple nearest-neighbor strategy, to derive label correction suggestions via multiple graph representations characterizing similar data manifolds. Hence, our method can take advantages of both nearest neighbors and manifold structures with the aid of graphs. 


The rest of this paper is organized as follows.
Some most relevant works are surveyed in Sec. \ref{sec:related}. Sec. \ref{sec:method} presents the proposed schemes for data splitting, multi-graph label propagation, and confidence-guided label correction. In Sec. \ref{sec:experiments}, experimental results on public datasets with noise labels are demonstrated. Finally, conclusions are drawn in Sec. \ref{sec:conclusion}.

 

%% file: TMM_sec02_review.tex
\subsection{Loss Functions}
\label{subsec:201}

To solve noisy label correction problem, several methods were developed by re-designing loss terms so that the learning system itself can re-weight the importance of training instances. 
For example, Wang et al. \cite{wang2017multiclass} employed an importance re-weighting strategy that enables the training on noisy data to reflect the results of noise free environment. Arazo et al. \cite{arazo2019unsupervised} evaluated their entropy-based per-sample loss for label correction by controlling the confidence of training sample via two weights. 
Still, some other losses, e.g., mean absolute error (MAE), generalized cross entropy (GCE), and reverse cross entropy (RCE), were proposed to avoid a biased learning system, the gradient saturation problem, or the overfitting/under-learning problem in the presence of noisy labels \cite{ghosh2017robust,zhang2018generalized,wang2019symmetric}. 
Nevertheless, although these methods prevent a learning system from fitting noisy samples, 
they may still be impracticable to complicated datasets due to the ignorance of hard samples.

\subsection{Label Correction}
\label{subsec:202}

Most recent label correction methods work based on i) a noise transition matrix, or ii) class prototypes.  
As for the former sort of methods, they are based on an instance-independent assumption to derive the transition probability, which depicts how possible a clean label flipping into a noisy one, independent to any type of data structure. 
However, this assumption cannot prevent the generation of incorrect labels, nor can it prevent the transition matrix from learning noisy labels. Therefore, Xia et al. \cite{xia2019anchor} proposed a transition revision method to address this issue, but their method is still limited by the fairness of the estimated initial transition matrix. 
Moreover, ``class prototype'' methods 
learn to represent each class via some class prototypes based on which data samples can receive corrected label information from their neighborhood \cite{sharma2020noiserank, han2019deep}. Nevertheless, these methods learn prototypes from a noisy dataset, so the quality and the accuracy of prototypes are still affected by the original noise distribution. 
Both these two types of methods learn from noisy data. 
Because there is no effective mechanism to verify the corrected labels suggested by the transition matrix or the class prototypes, the resulting pseudo-label may be incorrect and generate a worse correction suggestion if the class prototypes or the transition matrix overfits noisy training labels.

\subsection{Ensemble learning}
\label{subsec:203}

Some other methods exploit i) ensemble learning that feeds stochastically-augmented inputs to  a number of parallel sub-models to derive the corresponding prediction jointly with ii) an averaging method to derive label correction suggestions \cite{li2020dividemix, tarvainen2017mean}. These methods are based on the inconsistency among information learned by different sub-networks. For example, MLNT \cite{li2019learning} randomly changes the labels of each sub-network's training data to enhance the difference learned by each subnet, and DivideMix \cite{li2020dividemix} uses an additional regularization term to increase the prediction diversity among different sub-models. However, making sub-models learn to be different from each other cannot ensure the correctness of any sub-model itself, and they may be worse than they are supposed to be due to an unmatched noise distribution. 
For example, Yan et al.'s method is a method works only on uniformly distributed label noises \cite{yan2016robust}. In real-world applications, label noise usually distributes in some specific way rather than uniformly. Consequently, it is still intractable for simple ensemble learning approaches to filter out noisy labels.

\subsection{Label Propagation}
\label{subsec:204}

Label propagation is a graph-based semi-supervised method for generating pseudo-labels of unlabeled data based on given anchors \cite{zhou2004learning, iscen2019label, erdem2012graph}. 
%
Compared with the nearest-neighbor strategy that does not take the data manifold into account, label propagation methods can propagate labels from labeled anchors to unlabeled samples through a graph characterizing the manifold of the noisy dataset. 
For example, the method in \cite{douze2018low}
uses an affinity matrix to derive an approximate $k$-NN graph, through which it 
performs label propagation for a large-scale diffusion task.  Iscen et al. \cite{iscen2019label} proposed a 
transductive label propagation method to make label predictions by using i) a nearest-neighbor graph, ii) conjugate gradient based label propagation, and iii) a loss weighted by label certainty. 
Moreover, Bontonou et al. \cite{bontonou2019introducing} proposed a graph smoothness loss, maximizing the distance between samples of different classes, to derive better feature and manifold representations for classification purpose. 
This loss increases the robustness of a learning model against noisy training inputs, though its performance is still similar to that of cross-entropy because only distances between instances in different classes are considered.

%% file: TMM_sec03_method.tex

\subsection{Overview}

%

\begin{figure*}[t]
    \centering
    \includegraphics[width=0.9\textwidth]{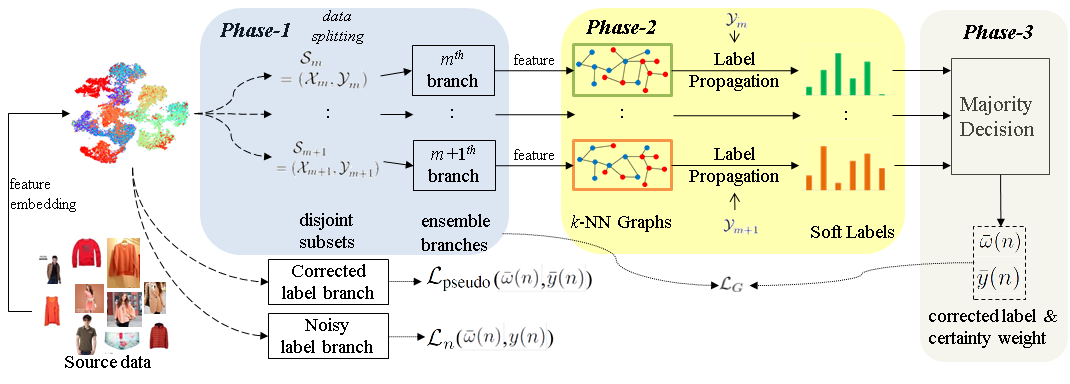}
    \caption{Framework of proposed ensemble-based label correction scheme. 
    The proposed method has primarily three branches, namely i) noisy label branch controlled by $\mathcal{L}_n$, ii) corrected label branch controlled by $\mathcal{L}_{pseudo}$, and iii) ensemble branches controlled by $\mathcal{L}_G$. Our design focuses on the ensemble branches implemented in three phases. The noisy label branch is used to perform \textit{feature embedding} and also used to regulate certainty weight jointly with the corrected label branch. Only the corrected label branch is used while deploying. The three phases of our ensemble branches are detailed in Sections \ref{subsec:302}--\ref{subsec:304}. Note that $y(n)$ denotes the original label given by the noisy dataset.
    }
    \label{fig:fig01}
    \vspace{-0.1in}
\end{figure*}

Fig. \ref{fig:fig01} illustrates the framework of the proposed method for label correction. Let $d_i = (x_i,y_i) \in \mathcal{D}$ denote a \textit{sample-label} pair from a data collection $(\mathcal{X}, \mathcal{Y})$, where $\mathcal{D} = \{ d_i  \}$ is the source noisy dataset.  Our ensemble learning scheme aims to split $\mathcal{D}$ into $M$ disjoint subsets, namely $\mathcal{S}_m$ for $m=1,\dots,M$, for training $M$ ensemble branches.
During the training, our method first trains $M$ classifiers $\Phi_m$ with parameter $\theta_m$, through the $M$ ensemble branches, independently on the $M$ disjoint subsets $\mathcal{S}_m$ derived by our data splitting strategy. Second, by feeding all training samples $x_i$ into each classifier, we extract totally $M$ different feature vector sets, each of which can be used to derive one graph representation of the data manifold of training dataset $\mathcal{X} = \{x_i\}$. Third, for given the $m$-th \textit{partial-label} set $(x^m_j, y^m_j) \in \mathcal{S}_m$ where only labels $y^m_j$ belonging to $\mathcal{S}_m$ are available, we construct its graph representation and use the graph to predict the label correction suggestions through label propagation for all data samples in the remaining $M-1$ partial-label sets. As a result, we use the $M$ partial-label sets to predict totally $M \times M$ label correction suggestions based on the $M$ graphs derived in the second step. Fourth, based on the corrected labels $\hat{y}^m_l \in \hat{\mathcal{Y}}_{m}$  and the $M$ \textit{sample-correction} sets $\{(x^m_l, \hat{y}^m_l) \, | \, \forall x^m_l \in \mathcal{S}_m\}$ obtained 
in the 
$l$-th training epoch, we generate another $M \times M $ label propagation suggestions. Finally, we derive the most likely correction result  $\bar{\mathcal{Y}} = \{\bar{y}_i\}$ from the total $2 \times M^2$ label correction suggestions via a majority decision. 
%

Specifically, based on the assumption that real-world noisy labels (e.g., obtained via crowd-sourcing) tend to be locally concentrated near decision borders, we devise a data splitting  strategy to partition the noisy source dataset $\mathcal{D}$ into $M$ disjoint subsets $\mathcal{S}_m$, each being the training set for one ensemble branch. 
In this way, $S_m$ is expected to retain the shapes of $k$ randomly-selected local neighborhoods, no matter noisy or not, as well as ensemble a random subset of $x_i \in \mathcal{X}$.  
Therefore, by taking a majority decision on the approximate manifolds learned by the $M$ sub-networks, the local influences brought by noisy samples can be mitigated. 
Finally, the \textit{noisy label branch} in our design is  trained on the whole original dataset $\mathcal{D}$ to derive the feature vectors of all $x_i$. 
\textbf{
Meanwhile, the \textit{corrected label branch} is trained based on corrected labels, i.e., $(\mathcal{X}, \bar{\mathcal{Y}})$, obtained after the \textit{l}-th training epoch, and it is the only branch used for deployment.
}

Our key idea is to restrict the influence of any group of  locally-concentrated noisy labels to only a minority of sub-networks via our local-patch-based data slitting strategy. As a result, most ensemble branches can learn their own relatively correct approximations of the data manifold around the noisy local patch, so that a suitable label correction result can be yielded by majority decision, accordingly. Also, no matter label noise is locally-concentrated or uniformly distributed in the data space, it can be effectively mitigated by the proposed ensemble method.
%
Overall, each training epoch of our method can be divided into three phases, as will be elaborated in Sec. \ref{subsec:302}--\ref{subsec:304}.

\subsection{Phase-1: Data Splitting/Re-splitting}
\label{subsec:302}

\begin{figure}
    \centering
    \includegraphics[width=0.45\textwidth]{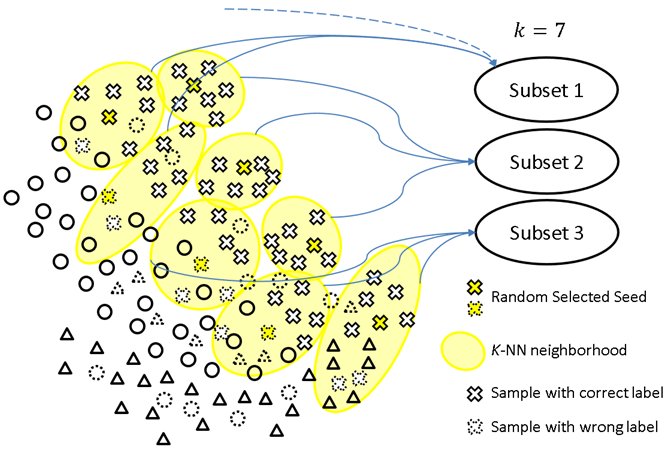}\\(a)\\
    \includegraphics[width=0.45\textwidth]{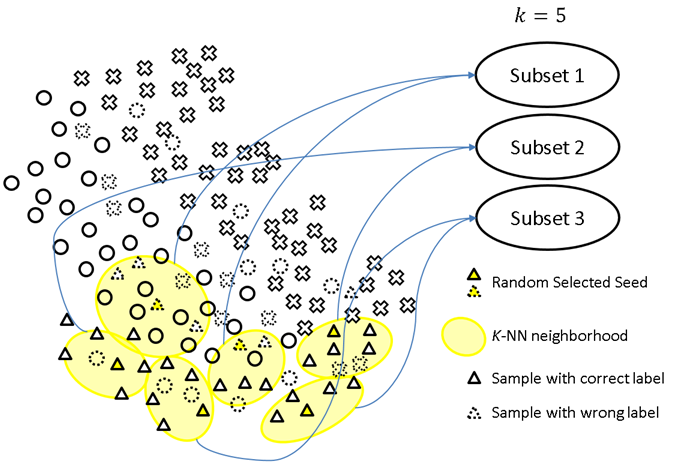}\\(b)
    \caption{Conceptual illustration of our data splitting strategy. 
    For each randomly selected seed (highlighted in yellow), we pick up only its \textbf{$k$-NN of the same class} (highlighted in white) to form a package, and $k=\frac{N_c}{M\dot B}$ with $N_c$ denoting the number of instances of class-$c$. This figure illustrates that each subset can be regarded as a random subset of the whole dataset if $B=1$. (a) An example of splitting ``crosses'' with $k=7$. (b) An example of splitting ``triangles'' with $k=5$. Note that \textit{dashed} samples are mis-annotated ones. 
    }
    \label{fig:fig02}
    \vspace{-0.05in}
\end{figure}

The first phase of each training epoch is  to randomly scatter per local neighborhood of data points on the source data manifold, including noisy labels, into disjoint subsets.
Specifically, as illustrated in Fig. \ref{fig:fig02}, for each selected $x_i$, we pick only its $k$ nearest-neighbors of the same class to form a \textit{local patch} around $x_i$ with $k = N_c / (M \times B)$, where $B$    is a predefined hyper-parameter, and $N_c$ denotes the cardinality of the class that $x_i$ belongs to. 
Therefore, because each disjoint subset $\mathcal{S}_m$ holds different local feature relationships of each class, we can confine the negative influence of each locally-concentrated noisy neighborhood to primarily one $\mathcal{S}_m$ and one ensemble branch by packaging every local $k$-nearest neighbors. This is the main advantage of the proposed method. 
%
As a result, on one hand, when a noisy local neighborhood only contaminates a minority of ensemble branches, our method can  vote down the negative influence of noisy labels by a majority decision. This case leads to a performance leap with our method:  the performance upper-bound. 
On the other hand, in the extreme case that all noisy labels are uniformly distributed globally, our data splitting strategy will not alter the noise distribution so that our ensemble model leads to a as good performance as typical random-selection schemes: the baseline performance. In sum, our method can be expected to provide a performance in between the upper-bound and the baseline. Since real-word label noise distribution tends to be concentrated on decision boundaries, our method can usually lead to performance improvements.

The proposed data splitting method is composed of the following steps: \\
\noindent \textbf{Initialization:} 
Determine the hyper-parameters, including the number of ensemble branches $M$ and the number of data packages per class per branch $B$. \\
\noindent \textbf{$k$-NN Packaging:} 
As illustrated in Fig. \ref{fig:fig02}, for each class, we separate its data into $M \cdot B$ packages, each containing a randomly selected seed sample and its $k$-NN in the feature domain with $k = \frac{N_c}{M \cdot B}$, where $N_c$ denotes the number of samples belonging to class $c$. 
Note that we adopt ``random sampling without replacement'' 
to guarantee that each sample and its $k$-NN can be assigned into only one package.\\
\noindent \textbf{Splitting:}
For each class, we randomly assign the $M \cdot B$ packages into $M$ disjoint subsets. Each subset $\mathcal{S}_m$ therefore stands for a coarse global approximation of the source data manifold 
but holds fine local manifold structures of different places  
independently.  


Moreover, we initialize each training epoch with the data splitting process, and we name this strategy \textbf{re-splitting}. 
Re-splitting enables each resemble branch to learn a different coarse approximation of the data manifold to prevent the resemble branches from overfitting and being biased to the same training data.
Because $\mathcal{S}_m$ is changed for each epoch, we need to initialize the model parameter $\theta^{l}_m$ for each ensemble branch to guarantee the fast convergence of the $l$-th training epoch by
\begin{equation}
   \theta_m^l = (1- \alpha_m) \theta_{\mbox{pseudo}}^{l-1} + \alpha_m \theta_{\mbox{noisy}}^{l-1} \mbox{,}
   \label{eq:updatetheta}
\end{equation}
where 
$\alpha_m^l$ is a random value within $[0, 1]$; and, $\theta_{\mbox{pseudo}}^{l-1}$ and $\theta_{\mbox{noisy}}^{l-1}$ are respectively the model parameters, obtained after the $(l-1)$-th training epoch, of the \textit{corrected-label branch} and that of the \textit{noisy-label branch}. 


\subsection{Phase-2: Multi-Graph Label Propagation}
\label{subsec:303}

We then construct the $m$-th graph representing the data manifold based on the features $f^m_i = f(x_i, \theta_m)$ extracted by the $m$-th model.  
 Then, for each graph, we can produce $M+M$ label correction suggestions by taking i) the original sample-label pair $(x_j^m, y_j^m) \in \mathcal{S}_m$ and ii) the sample-correction pair $(x_j^m, \hat{y}_j^{m,l})$ obtained in the $l$-th training epoch as different starting partial-labels, where $\bar{y}_j^{m,l}$ denotes the $j$-th sample's label correction suggestion given by the $m$-th ensemble branch in the $l$-th training epoch.
As a result, our method can hence offer totally $M \cdot (M+M)=2M^2$ sets of label correction suggestions.

%

Our multi-graph label propagation strategy works based on the features extracted by classification models $\Phi_m(\cdot; \theta_m)$ of the $M$ ensemble branches. The first step is to build $M$ normalized weighted undirectional adjacency matrices $\mathcal{W}_m$, each  representing the $m$-th graph constructed based on $f_i^m$, for the label propagation process described in (\ref{eq:labelprop}), and $\mathcal{W}_m$ is required to be symmetric \cite{zhou2004learning}.
The matrix $\mathcal{W}_m$ is defined as 
\begin{eqnarray}
\mathcal{W}_m &=& D_m^{-\frac{1}{2}} (A_m + A^T_m) D_m^{-\frac{1}{2}} \mbox{,}
\label{eq:04}
\end{eqnarray}
where $D_m$ is a diagonal matrix used to normalize $(A_m + A^T_m)$, and $A_m$ is a  directional adjacency matrix whose ($s$,$t$)-th entry is defined as follows: 
\begin{equation}
    A_m(s, t) = \bigg\{ 
    \begin{array}{ll}
         sim( f^m_{s}, f^m_{t})^\gamma &
         \mbox{, for } f^m_{s} \in \mathcal{N}(f^m_{t}), s\neq t\\
         0 & 
         \mbox{, otherwise}
         \end{array}
\label{eq:eq03}
\end{equation}
where $f^m_{s}$ is the feature of the $s$-th sample extracted by the model of the $m$-th branch,  
$\mathcal{N}(f^m_{t})$ denotes the collection of the $k$ nearest neighbors of $f^m_{t}$, $sim(\cdot, \cdot)$ denotes the cosine similarity metric ranging in $[0,1]$, and $\gamma$ is a predefined hyper-parameter. Note that $A_m$ is asymmetric.

Next, we propagate the label information of i) the sample-label pairs $(x^m_j, y^m_j)$ in the $m$-th disjoint subset $\mathcal{S}_m$ and ii) sample-correct pairs $(x^m_j, \bar{y}^{m,l}_j)$ in turn through each of the $M$ graphs to obtain totally $2M^2$ label correction suggestions, i.e., partial-labels, for each data sample $x_i$. 
We adopt such a strategy because of the following two concerns. First, while using only original noisy labels $y^m_j$ to derive the label propagation result, 
the obtained labels are probably contaminated, so this setting surely limits the accuracy of obtained labels. 
Second, while using only corrected labels $\hat{y}^{m,l}_j$, the propagation result may overfit the corrected labels, thereby misleading the label correction process. Therefore, we exploit both the original noisy labels $y^m_j$ and corrected labels $\hat{y}^{m,l}_j$ in our label propagation process.

Let $Z_{m,j}$ denote the label correction suggestion, the label propagation process \cite{zhou2004learning} is conducted by
\begin{equation}
    Z_{m,j} = (\mathbf{I} - \alpha \mathcal{W}_m)^{-1} Y_j \mbox{,}
    \label{eq:labelprop}
\end{equation}
where $Y_j$ is an $N \times C \times 2$ matrix that records the reference labels associating the data samples belonging to the $j$-th disjoint subset $\mathcal{S}_j$, $\mathcal{W}_m$ is the normalized weighted undirectional adjacency matrix defined in (\ref{eq:04}), and $N$ and $C$ denote respectively the cardinality of the original dataset $\mathcal{X}$ and the cardinality of $\mathcal{Y}$, i.e., the number of total classes. 
Note that the third dimension of $Y_j$ contains two partial-labels: the original noisy one  $y^m_j$ and the corrected one $\hat{y}^{m,l}_j$. 

Finally, the partial-label suggested by the $m$-th ensemble branch based on the label information 
of 
$\mathcal{S}_j$ becomes
\begin{equation}
    \hat{Y}_{m,j}(n,:) = \mathop{\arg\!\max_{c}} Z_{m,j}(n,c,:) \mbox{,}
\label{eq:eq08}
\end{equation}
where every $Z_{m,j}$ is an $N \times C \times 2$ matrix, and  $Z_{m,j}(n, c, \cdot)$ indicates how possible the $n$-th sample belongs to the $c$-th class. 
After obtaining $2M^2$ different label suggestions $\hat{Y}_{m,j}$, it turns to Phase-3 to derive the final correction result.

\subsection{Phase-3: Confidence-Guided Label Correction}
\label{subsec:304}

We select the most frequent category among all label suggestions $\hat{Y}_{m,j}(n,:)$ via a majority decision as the final label correction result $\bar{y}(n)$.

To guide the label correction, we propose a normalized average confidence level to devise our loss function. 
The normalized average confidence level $\bar{\omega}(n)$ of the final label correction result $\bar{y}(n)$ is defined as 
\begin{equation}
\bar{\omega}(n) = \frac{\hat{\omega}(n) - \min\{\hat{\omega}(n)\}}
{\max\{\hat{\omega}(n)\} -\min\{\hat{\omega}(n)\}} \mbox{,}
\label{eq:eq12}
\end{equation}
where
\begin{equation}
\hat{\omega}(n) = \frac{1}{2M^2}
\sum_{(m,j,q) \in \Omega_n } \omega_{m,j}(n,q) \mbox{,}
\label{eq:eq13}
\end{equation}
where $\Omega_n = \{ (m,j,q) \mid \bar{y}(n) = \hat{Y}_{m,j}(n, q)\}$ means that only those samples, whose label correction suggestions $\hat{Y}_{m,j}(n, q)$ are identical to the final corrected label $\bar{y}(n)$, are used to calculate the normalized average confidence level.
Therefore, $\bar{\omega}$ can be maximized 
when i) the predictions have a high confidence level, which implies a relatively low entropy, or ii) the $n$-th corrected labels suggested by all graphs are of the same category. 

Note that the confidence level $\omega_{m,j}(n)$
of the label propagation result $Z_{m,j}$ can be assessed based on the certainty weight described in \cite{iscen2019label}: 
\begin{equation}
\omega_{m,j}(n, :) = 1 - \frac{H(\bar{Z}_{m,j}(n,c,:))}{\log(C)} \mbox{,} 
\label{eq:eq09}
\end{equation}
where
\begin{eqnarray}
\bar{Z}_{m,j}(n,c,:) &=& \frac{Z_{m,j}(n,c,:)}{\sum_k Z_{m,j}(n,k,:)} \mbox{,}
\nonumber
\\
H(\bar{Z}_{m,j}(n,c,:)) &=& -\sum_{c=1}^C \bar{Z}_{m,j}(n,c,:) \log(\bar{Z}_{m,j}(n,c,:)) \mbox{,}
\nonumber
\end{eqnarray}
where $\bar{Z}_{m,j}$ records the occurrence frequency of each label in the label propagation result matrix $Z_{m,j}$, and $H(\cdot)$ denotes the entropy. 
Therefore, a larger $\omega_{m,j}(n, q)$ implies a label $\hat{Y}_{m,j}(n,q)$ with higher confidence (i.e., lower uncertainty). 

After normalizing the final weights described in (\ref{eq:eq12}), we use i) the corrected labels and ii) the weighted cross entropy loss to train the next-epoch ensemble branches and the next-epoch corrected label branch. The loss functions used to train our ensemble model are detailed in the next subsection.


\subsection{Loss Functions}
\label{subsec:305}

We adopt  the following weighted cross-entropy to measure the loss of corrected (pseudo) labels branch:
\begin{equation}
    \mathcal{L}_{\mbox{pseudo}} = - \sum_{n} \bar{\omega}(n) \bar{y}(n) \log p_{\mbox{pseudo}}(n, \bar{y}(n)) \mbox{,}
    \label{eq:eq15}
\end{equation}
where $\bar{y}(n)$ is the pseudo label of $x(n)$ derived from our label correction process in the current training epoch, and $p_{\mbox{pseudo}}(n, \bar{y}(n))$ denotes the  probability, predicted by the softmax layer of the corrected (pseudo) label branch, 
of $x(n)$ belonging to class $\bar{y}(n)$.

Then, the noisy label branch is trained via the weighted cross-entropy loss, i.e.,  
\begin{equation}
\mathcal{L}_{\mbox{noisy}} = - \sum_n \ell_{\mbox{noisy}}(n)\mbox{,}
\label{eq:ce_loss}
\end{equation}
where
\begin{eqnarray}
\ell_{\mbox{noisy}}(n) &=& \bigg \{
\begin{array}{ll}
    \bar{\omega}(n)  y(n)  \log p_{\mbox{noisy}}(n, c) & \mbox{, if }c=\bar{y}(n) \\
    (1-\bar{\omega}(n))  y(n)  \log p_{\mbox{noisy}}(n, c) & \mbox{, otherwise,} \nonumber
    \label{eq:eq16}
\end{array}
\end{eqnarray}
where $p_{\mbox{noisy}}$ denotes the probability predicted by the softmax layer of the noisy label branch. 
This loss means that if the \textit{n}-th sample 
is corrected, 
we give it a complementary confidence level $1 - w(n)$ to highlight 
its low confidence. 

Moreover, because the $M$ ensemble branches are mainly designed to construct the $M$ graphs rather than predicting the classification probability of each sample, we exploit a graph smoothness loss modified from \cite{bontonou2019introducing} to 
derive a set of graph embedding via which the inter-class distances are maximized, similar to the concept described in \cite{wang2020graph}.
This loss uses radial basis functions to measure the weighted distance between the features of any two samples of different classes. 
Because the two samples may come from different $\mathcal{S}_m$, this loss can lead to a large enough between-class margin on all graphs. That is,
\begin{equation}
    \mathcal{L}_G = \sum_{(s,t)}
    \sqrt{\bar{\omega}_s \bar{\omega}_t } \exp \big\{ -\alpha \| \mathbf{p}^{\mathcal{I}(s)}_s - \mathbf{p}^{\mathcal{I}(t)}_t \|_2 \big\} \mbox{,} 
    \label{eq:eq17x}
\end{equation}
where $(s,t) \in \Omega= \{(s, t) \mid \bar{y}(s) \neq \bar{y}(t) \}$, $\mathcal{I}(s)$ is an indicator function that returns the index of the disjoint subset containing the $s$-th sample, and $\mathbf{p}^{\mathcal{I}(s)}_s$ denotes the $s$-th sample probability vector generated by the softmax layer of the subnetwork trained on the $\mathcal{I}(s)$-th subset.
This loss forces the distance between two nodes of different classes on the graph to be large.

$\mathcal{L}_G$ has an additional weighting factor $\sqrt{\bar{w}_s \bar{w}_t}$ which gives an instance pair with a lower confidence level a lower weight to alleviate possible negative effects of instances with lower confidence levels. 
$\mathcal{L}_G$ is different from typical entropy loss and triplet loss \cite{schroff2015facenet} designs because $\mathcal{L}_G$ aims to map instances of different classes into feature clusters far away from each other. On the contrary, both cross-entropy loss and triplet loss encourage the network to map instances of the same class into the similar features. Consequently, $\mathcal{L}_G$ can assist our model to produce distinguishable features for label correction tasks. 
Finally, the total loss of the proposed model is 
\begin{equation}
    \mathcal{L}_{\mbox{total}} = \mathcal{L}_G + \mathcal{L}_{\mbox{pseudo}} + \mathcal{L}_{n} \mbox{.}
    \label{eq:eq19}
\end{equation}

%% file: TMM_sec04_experiment.tex
\subsection{Datasets}
\label{subsec:401}

We evaluate the proposed method on i) simulated noisy datasets based on CIFAR-10/100 and ii) two other real-world noisy datasets: \textbf{Clothing1M} \cite{xiao2015learning} and \textbf{Food101-N} \cite{lee2018cleannet}. The properties of Clothing1M and Food101-N are described below. 

First, Clothing1M contains images with 14 categories of fashion clothes. The data in Clothing1M are collected from the Internet, and their category labels are generated based on the surrounding texts, which leads to a label noise rate of 38.46\%. Clothing1M also provides a small clean set with labels corrected by human annotators. The numbers of images in Clothing1M's label-corrected training, validation, testing sets, and the noisy labeled set are about 47.5k, 14.3k, 10.5k, and 1M, respectively. In the four sets, only the 1M noisy labeled set contains noisy labels.
We use the noisy labeled set to train our model to evaluate its performance, and then we conduct the ablation study jointly on the noisy labeled set and the human-annotated clean testing set. 

Second, Food101-N contains about 310K food images of 101 categories. The images in Food101-N are collected from the Internet with a noisy label rate of 20\%. 
To evaluate the performance of our method, we apply it on Food101-N and derive a label correction result first, and then we train a classifier on the label-corrected dataset and conduct testing on Food-101's testing set \cite{bossard2014food}. 
Note that Food-101's testing set contains only 55k images with clean labels. 


\subsection{Experiment Settings}

We summarize our experiment settings as the following four points.  
First, after the data splitting phase of each training epoch, we resize training images $x^m_i \in \mathcal{S}_m$ to $256 \times 256$, and then cropped them randomly into $224 \times 224$ patches. Hence, training samples for each training epoch are different variants of the source data, which improves the diversity of training samples for each epoch.
Second, the optimizer is SGD, and the momentum value is $0.9$. The initial learning rate is $0.01$, and it is reduced by a factor of $\frac{1}{10}$ every $5$ epochs. The maximal training epoch is set to be $15$, and the batch size is $64$. We also adopt ridge (L2) regularization that is weighted by $5\times 10^{-3}$. 
Third, the number of branches is $M=5$, and  the  number  of  data  packages per class in each branch is $B=4$.
Fourth, the features of all training samples are extracted \textit{initially} by a ResNet-50 \cite{he2016deep} pretrained on ImageNet. After the first training epoch, the features used for our data (re-)splitting phase are derived by the latest network model trained in the noisy label branch.

\subsection{Experiments on Clothing1M}

We evaluate our model's performance on Clothing1M through two different scenarios. 
The first scenario is to train our model and other state-of-the-arts on only noisy datasets without any additional supervision. 
In the second scenario, all models are trained on the same hybrid set consisting of i) the noisy set and ii) the clean set consisting of 50K human-annotated images, and then the obtained models are further fine-tuned on the 50K-image clean dataset via the cross-entropy loss. The results of the first and the second parts are shown respectively in Tables \ref{tab:newt1t2a} and \ref{tab:newt1t2b}. Note that in the second scenario, the labels of the clean dataset are kept unchanged, and their certainty weight are set to be $1$. 
Table \ref{tab:newt1t2a} demonstrates that our method achieves state-of-the-art performance, i.e., 75.17\% accuracy, compared with the other methods. Meanwhile, in the second scenario, our method still outperforms all compared state-of-the-arts without any fine-tuning and is comparable to Self-Learning \cite{han2019deep} after fine-tuning with the 50k clean set.

\begin{table}{t}
	\caption{\small Classification accuracy of different methods on the noisy set of Clothing1M 
	}
	\centering
\begin{tabular}{|l l |}
\hline
Method  &  Acc.(\%) \\
\hline
CrossEntropy &  69.21 \\
MLNT~\cite{li2019learning} &  73.47 \\
T-Revision~\cite{xia2019anchor} & 74.18 \\
Self-Learning~\cite{han2019deep} & 74.45 \\
MetaCleaner~\cite{zhang2019metacleaner} & 72.50 \\
DivideMix~\cite{li2020dividemix} & 74.76 \\
NoiseRank~\cite{sharma2020noiserank} & 73.82 \\
ELR~\cite{liu2020early}& \textcolor{red}{74.81}  \\
\hline
Ours & \textbf{75.17} \\
\hline
\end{tabular}
\label{tab:newt1t2a}
\end{table}

\begin{table}{h}
	\caption{\small Classification accuracy of different methods on a hybrid set of the noisy set and the clean set, containing 50K images, of Clothing1M 
	}
	\centering
\begin{tabular}{| l c c|}
\hline
Method  & Acc.(\%) & Fine-tuning\\
\hline
CrossEntropy & 75.67 & x\\
DataCoef~\cite{zhang2020distilling} & \textcolor{red}{77.21} & x \\
\textbf{Ours} & \textbf{78.01} & x\\
\hline
CrossEntropy & 80.32 & o \\
Forward~\cite{patrini2017making} & 80.38 & o \\
MetaCleaner~\cite{zhang2019metacleaner} & 80.78 & o \\
CleanNet~\cite{lee2018cleannet} & 79.90 & o\\
Self-Learning~\cite{han2019deep} & \textbf{81.16} & o \\
\textbf{Ours} & \textcolor{red}{81.13} & o  \\
\hline
\end{tabular}
\label{tab:newt1t2b}
\end{table}

\subsection{Experiments on Food101-N}
\label{subsec:404}

\begin{figure}[t!]
    \centering
    \includegraphics[width=0.48\textwidth]{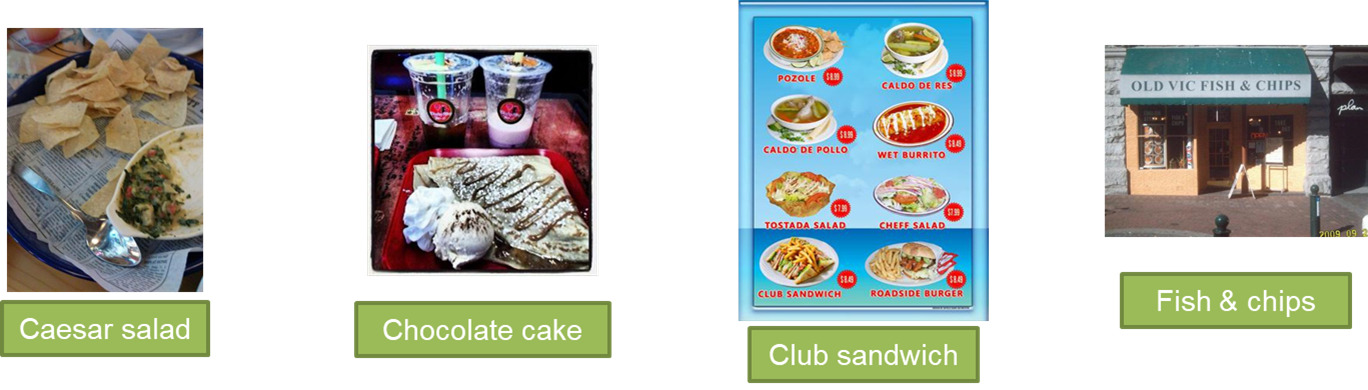}
    \caption{Samples of three multi-label images and one anomaly in Food101-N. 
    }
    \label{fig:fig03}
\end{figure}

The classifier, trained on the data corrected by our method, achieves an second best accuracy of 85.13\% on Food101-N, comparable to the best performance of 85.79\% achieved by NoiseRank \cite{sharma2020noiserank}, as demonstrated in Table \ref{tab:newt3food}. 
Although the noise rate of Food101-N is about 20\%, much lower than that of Clothing1M, the noisy labels in Food101-N are much more complicated. 
Incorrect labels in Food101-N involve three different types: i) photos with confusing contents, ii) multi-label images, e.g. photos with multiple foods, and iii) anomalies, e.g. a shopfront photo labeled as ``fish and chips". Fig. \ref{fig:fig03} shows four samples of the second and the third types of mis-annotations. 
Because our method is designed for confusing label noises, i.e., only one of the three types of mis-annotations in Food101-N, the performance of our method is just comparable with typical data cleaning methods like NoiseRank \cite{sharma2020noiserank}. 
Nevertheless, data cleaning methods tend to remove hard samples simultaneously, which leads to irreversible information loss.  Consequently, our method works much better than NoiseRank on Clothing1M, a dataset containing no anomalies, but not so good as NoiseRank in fighting against multi-label samples and anomalies in Food101-N.

\begin{table}[t]
\caption{\small Classification accuracy of different methods on Food101-N dataset with additional 50k clean set.}
\begin{center}
\begin{tabular}{|l r|}
\hline
Method  & Acc.(\%) \\
\hline
CrossEntropy & 81.97 \\
CleanNet~\cite{lee2018cleannet} & 83.95 \\
MetaCleaner~\cite{zhang2019metacleaner} & 85.05 \\
Self-Learning~\cite{han2019deep} & 85.11 \\
NoiseRank~\cite{sharma2020noiserank} & \textbf{85.79}\\
\textbf{Ours} & \textcolor{red}{85.13} \\
\hline
\end{tabular}
\end{center}
\label{tab:newt3food}
\end{table}

\subsection{Analyses and Discussions}
\label{subsec:406}

In this subsection, we discuss the effect of each component in our method, including i) the number of local manifolds $B$, ii) the number of ensemble branches $M$, iii) the settings of $\mathcal{L}_G$, and iv) the re-splitting process.

\begin{figure}[t]
    \centering
    \includegraphics[width=0.48\textwidth]{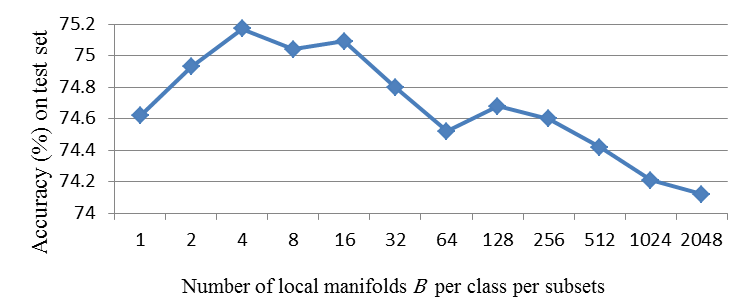}
    \caption{Test accuracy (\%) for different numbers of local manifolds ($B$) with $M=5$.}
    \label{fig:fig04}
    \vspace{-0.25in}
\end{figure}

\noindent $\bullet$ \textbf{Number of local manifolds} $B$\\
\indent 
Fig. \ref{fig:fig04} shows that when $M=5$, our method achieves the best accuracy 75.17\% on the noisy set of \textbf{Clothing1M} when $B=4$. In addition, $B=1$ and $B=2,048$ are two extreme conditions,. The former represents that each disjoint subset $\mathcal{S}_m$ contains one large local patch, and therefore it is impossible for each subset to approximate the global data manifold. On the contrary, the latter denotes a strategy similar to random sampling, implying that each subset can coarsely approximate the whole data manifold yet loses the ability to retain the local information. When $B=2,048$, the proposed ensemble learning strategy has no advantage in coping with noisy labels because data in all disjoint subsets follow the same distribution.

\begin{figure}[t]
    \centering
    \includegraphics[width=0.48\textwidth]{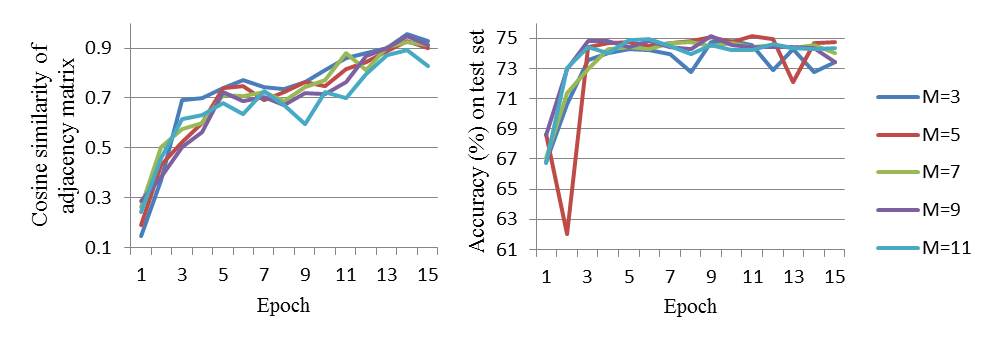}
    \caption{(a) Average cosine similarity in each adjacency matrix with epoch from 1 to 15 in training process. (b) Test accuracy (\%) with epoch from 1 to 15 in training process. The line denote different setting of $M$ and all the model use same $B=4$.}
    \label{fig:fig05}
    \vspace{-0.20in}
\end{figure}

\noindent $\bullet$ \textbf{Number of Ensemble Branches} $M$\\
\indent 
The number of ensemble branches, $M$, is also a key hyper-parameter in our method. 
To evaluate the impact of $M$, we conduct experiments by fixing $B=4$. As shown in Fig. \ref{fig:fig05}(a), the ensemble branches are beneficial for estimating data manifold because the cosine similarities in the adjacency matrices of derived graphs roughly increases with the number of ensemble branches. However, Fig. \ref{fig:fig05}(b) shows that the testing accuracy tends to be saturated when $M \geq 5$. This implies that when $M \geq 5$, some ensemble branches may learn nearly the same coarse approximation of data manifold, and such information might be redundant thus cannot further improve the generalization performance. Therefore, we set $B=4$ by default. 

Moreover, Fig. \ref{fig:fig05} provides another interpretation. By checking the average cosine similarities in an adjacency matrices derived by different ensemble branches during training, we find that the final ensemble model performs the best when the average cosine similarity is in between $0.7$ and $0.8$. When the average cosine similarity reaches $0.9$, the model becomes overfit because the testing accuracy of the ensemble model decreases, slightly though.

\begin{figure}[t]
    \centering
    \includegraphics[width=0.48\textwidth]{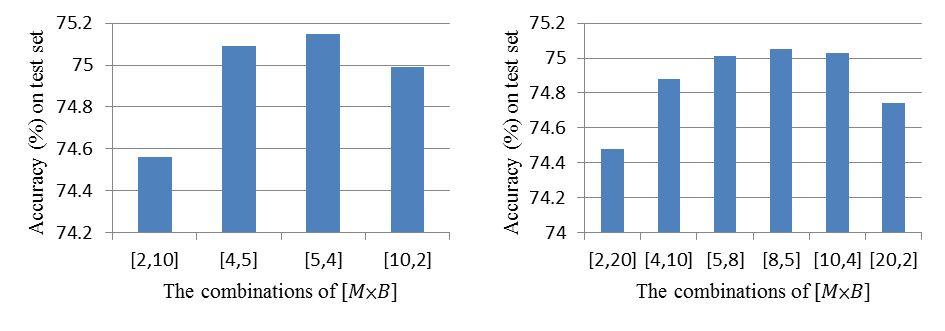}
    \caption{(a) Test accuracy (\%) with different combinations of $[M,B]$ and $M B = 20$. (b) Test accuracy (\%) with different combinations of $[M,B]$ and $M B=40$.}
    \label{fig:fig06}
    \vspace{-0.25in}
\end{figure}

\noindent $\bullet$ \textbf{Effects of different combinations of $(M, B)$}\\
\indent 
Fixing the number of local patches per class, i.e., $M \cdot B$, we here discuss the influence brought by different combinations of $M$ and $B$.
The experiments demonstrated in Fig. \ref{fig:fig06} are derived by setting $M \cdot B=40$ and $M \cdot B = 20$. 
For $M \times B = 20$, we conduct experiments on $(M, B) = (2, 10)$, $(4,5)$, $(5,4)$, and $(10,2)$. As for $M \times B = 40$, we examined $(M, B) = (2, 20)$, $(4,10)$, $(5,4)$, and $(10,2)$ in turn. 
Based on these results, we have two observations.
First, $(M, B) = (4,5)$ achieves the best accuracy of $75.15\%$, whereas $(M, B) = (8,5)$ reaches about $75.05\%$. Second, $M$ is far more critical than $B$, and a suitable $M$ should range in between 4 and 10. This is because once a too large $M$ is used, the source noisy data will be divided into too small disjoint subsets, and such small subsets cannot retain the shape of the entire data manifold. 

\noindent $\bullet$ \textbf{Weighted Graph Smoothness Loss}\\
\indent 
Our $\mathcal{L}_G$ described in (\ref{eq:eq17x}) is measured by the $L_2$ distance between the probability vectors $\mathbf{p}_s^{\mathcal{I}(s)}$ produced by the softmax layer of different ensemble branches. This design encourages different sub-networks to produce the same classification results instead of similar feature vectors. To verify the effectiveness, we run experiments via JS-divergence, KL-divergence, and feature vectors for comparisons, as shown in Table \ref{tab:newT05}. First, the experiments show that $\mathcal{L}_G$ is more effective than cross-entropy loss because cross entropy tends to make all instances in the same class have features of the same kind but $\mathcal{L}_G$ does not. Second, by simply using the $L_2$ distance between probability vectors, i.e., GS+$p$+L2, rather than the $L_2$ distance between feature vectors, i.e., GS+$f$+L2, the accuracy of our model can be boosted from 74.47\% to 75.15\%. Third, no matter $L_2$ distance, KL-divergence, or JS-divergence is adopted, the softmax output can consistently boost the model accuracy up to 75.17\%. This proves the effectiveness of $\mathcal{L}_G$ and the usage of softmax output. Note that wGS denotes \textit{weighted graph smoothness loss}, $p$ probability vector, and $f$ feature vector.




\begin{table}[h]
\begin{center}
	\caption{\small Classification accuracy of different GRAPH SMOOTHNESS LOSS (GS) on Clothing1M.}
\begin{tabular}{|l r|}
\hline
\small
Method  & Acc.(\%) \\
\hline
\small
CrossEntropy & 74.08 \\
GS + Feature + L2 & 74.47 \\
GS + Softmax + L2 & 75.15 \\
Weighted GS + Softmax + L2 & \textbf{75.17} \\
Weighted GS + Softmax + KL & 74.92 \\
Weighted GS + Softmax + JS & 74.98 \\
\hline
\end{tabular}
\end{center}
\vspace{-0.11in}
\label{tab:newT05}
\end{table}

\noindent $\bullet$ \textbf{Data Re-splitting}\\
\indent 
Our experiment shows that the data re-splitting process described in Sec. \ref{subsec:302} can on average increase the accuracy by about 0.04\%. Specifically, without data re-splitting the accuracy is on average 75.13\%, whereas the accuracy on average increases to 75.17\% with the aid of data re-splitting.
This experiment supports the idea that re-splitting can assist ensemble branches to learn different combination of local patches and also prevent them from overfitting. 

We validate the data re-splitting process in the following two steps. First, we extract the feature vectors of images in the to-be-corrected noisy dataset by using a ResNet-50 pretrained on Clothing1M via cross entropy.
Second, with the obtained features, we split the to-be-corrected noisy dataset into $M$ disjoint subsets and then apply our method for label correction without data re-splitting.


%

\noindent $\bullet$ \textbf{Certainty weight} 

\begin{figure*}
    \centering
    \includegraphics[width=0.9\textwidth]{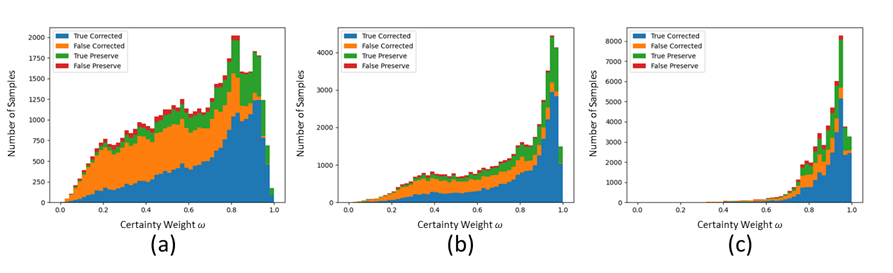}
    \caption{Distribution of corrected labels and certainty weights on Cifar-10 with 80\% symmetric noise at (a) the 10-th epoch, (b) the 100-th epoch, and (c) the 200-th epoch. Note that the horizontal axis denotes the value of certainty weight, and the vertical axis represents the data amount.}
    \label{fig:s1}
\end{figure*}

Fig. \ref{fig:s1} illustrates how certainty weight varies with training epochs. The certainty weight described in (\ref{eq:eq12}) represents the probability of the correctness of a corrected label. These three figures jointly show that our system design is valid because the number of samples with larger certainty weights increases with the number of  training epochs.

\subsection{Ablation Study}
\label{subsec:405}

\begin{figure*}
    \centering
    \includegraphics[width=0.75\textwidth]{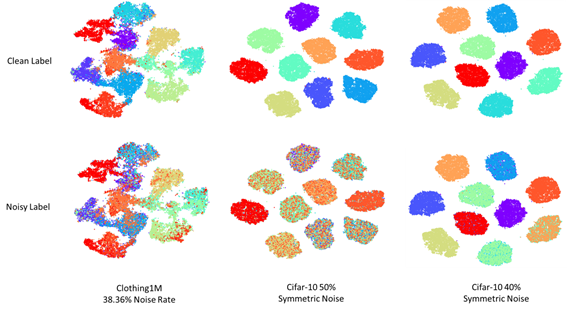}
    \caption{t-SNE maps of clean training datasets (the top row) and to-be-corrected noisy datasets (the bottom row). Left: Clothing1M with 38.36\% noise rate. Middle: CIFAR-10 with 50\% symmetric noise. Right: CIFAR-10 with 40\% symmetric noise.}
    \label{fig:s0}
\end{figure*}

In this section, we examine our system architecture design and clarify the effects brought by different combinations of loss terms and system components for all variants of the proposed method.  
We first examine the performance of each of our model variants on CIFAR-10 with man-made symmetric noises. Fig. \ref{fig:s0} shows the t-SNE maps of datasets used for this part of ablation study. 

In this experiment set, what we examine include i) the necessity of  noisy label branch in our default architecture, ii) the impact of  Mixup \cite{zhang2017mixup} used for data augmentation, and iii) the effects brought by different loss terms used in the ensemble branches.
Mixup is a procedure developed for data augmentation for semi-supervised learning, that enables the obtained model to produce linear prediction results between any pair of samples. Hence, it can yield smooth and robust pseudo labels and also avoid overfitting problems. 
We summarize the comparison among different loss term designs as follows: \\
\\\noindent $\bullet$ \textbf{Ensemble branch:} \\
Although the graph smoothness loss described in (\ref{eq:eq17x}) is beneficial for deriving a better adjacency matrix, this loss may overfit wrongly-labelled samples easily. Hence, in our default setting, we measure this loss only on those samples whose labels remain unchanged after our correction procedure. However, such setting may reduces the effectiveness of this loss, because, when the loss rate is too high, the amount of uncorrected samples participating in the computation of the graph smoothness loss would be too small. 
As a result, we measure the weighted sum of i) the \textbf{MAE} (mean absolute error) $\mathcal{L}_{en\_mae}$ between $1$ and the classification probability of corrected samples, and ii) the \textbf{cross entropy} $\mathcal{L}_{en\_ce}$ of uncorrected samples as an alternative of the loss term for ensemble branches. We then compare this alternative design with our default setting $\mathcal{L}_G$ described in (\ref{eq:eq17x}). \\
\noindent $\bullet$ \textbf{Corrected label branch:} \\
Similarly, we adopt the sum of i) \textbf{MAE} $\mathcal{L}_{cor\_mae}$ and ii) \textbf{cross entropy} $\mathcal{L}_{cor\_ce}$ as an alternative loss design for the corrected label branch. We use the soft-label, which is estimated as a convex combination, weighted by the certainty weight $\bar{\omega}$ depicted in (\ref{eq:eq12}), of the prediction result derived by the corrected label branch and that derived the ensemble branches, to compute both the $\mathcal{L}_{cor\_mae}$ and the $\mathcal{L}_{cor\_ce}$. A larger $\bar{\omega}$ will produce a soft-label closer to the label suggested by the corrected label branch. 
In addition, the weighting factors for loss terms used in our ablation study are listed in Table \ref{fig:s2}.

\begin{table}[]
    \caption{Noise rate versus weighting factors for alternative loss terms used in our ablation study. \textbf{(This table needs re-layout.)}}
    \centering
    \begin{tabular}{
    |c|c|c|c|c|c|}
    \hline 
         & \multicolumn{5}{|c|}{Cifar-10} \\ \hline
         Noise Rate & 20 & 50 & 80 & 90 & asym 40\%  \\ \hline
         $\mathcal{L}_{en\_ce}$ & 1 & 1 & 1 & 1 & 1 \\ \hline
         $\mathcal{L}_{en\_mae}$ & 0 & 0 & 1 & 1 & 0 \\ \hline
         $\mathcal{L}_{cor\_ce}$ & 1 & 1 & 1 & 1 & 1 \\ \hline
         $\mathcal{L}_{cor\_mse}$ & 0 & 1 & 10 & 10 & 1 \\ \hline\hline
          & \multicolumn{4}{|c|}{Cifar-100} & Clothing1M \\ \cline{1-5}
         Noise Rate & 20 & 50 & 80 & 90 & \\ \hline
         $\mathcal{L}_{en\_ce}$ & 2.5 & 2.5 & 2.5 & 2.5 & 1 \\ \hline
         $\mathcal{L}_{en\_mae}$ & 0 & 1 & 1 & 1 & 1 \\ \hline
         $\mathcal{L}_{cor\_ce}$ & 1 & 1 & 1 & 1 & 1 \\ \hline
         $\mathcal{L}_{cor\_mse}$ & 10 & 40 & 80 & 80 & 1 \\ \hline
    \end{tabular}
    \label{fig:s2}
\end{table}

\begin{table}[]
    \centering
    \caption{Test Accuracy (\%) of experiments on Clothing1M. Each row corresponds a different system architecture.}
    \begin{tabular}{|l|r|}
    \hline
              & Acc (\%) \\ \hline
    Ours (default)         & 75.17 \\ \hline
    Ours, w/o noisy branch         & 73.91 \\ \hline
    Ours + Mixup, w/o noisy branch & 73.87 \\ \hline
    Ours + Mixup + Soft Label, w/o noisy branch & 73.90 \\ \hline
    \end{tabular}
    \label{fig:s6}
\end{table}

Finally, Tables \ref{fig:s3}, \ref{fig:s5}, and \ref{fig:s6} show that by removing the noisy label branch, by changing the loss functions, or by using soft-labels, the proposed system can become suitable for correcting CIFAR-10/CIFAR-100 datasets with man-made random label noises. However, such modifications degrade our system performance for real-world applications, e.g. Clothing1M, whose data are contaminated by confusing label noises, as shown in Fig. \ref{fig:s6}. Hence, we finally choose to adopt our default architecture and loss settings because this combination is most beneficial for real-world applications.

\begin{table}[t]
    \centering
    \caption{Test Accuracy (\%) of experiments on Cifar-10 with symmetric and asymmetric noises. For symmetric noises, the note rate was set to be 20\%, 50\%, 80\%, and 90\% in turn, whereas the noise rate of the experiment on asymmetric noise is fixed to be 40\%.
    Note that values in \textbf{Last} rows are the mean test accuracy values derived by models of last 10 training epochs. }
    \begin{tabular}{|l|l|r|r|r|r|r|}
    \hline
    \multicolumn{2}{|r|}{Noise Level (\%)}
                  & 20 & 50 & 80 & 90 & \small asym\\ \hline
    \small Cross Entropy & \small Best & \small 86.8 & \small 79.4 & \small 62.9 & \small 42.7 & \small 85.0\\
                  & \small Last & \small 82.7 & \small 57.9 & \small 26.1 & \small 16.8 & \small 72.3\\ \hline
    \small Mixup \cite{zhang2017mixup} & \small Best & \small 95.6 & \small 87.1 & \small 71.6 & \small 52.2 & --\\
                  & \small Last & \small 92.3 & \small 77.6 & \small 46.7 & \small 43.9 & --\\ \hline 
    \small P-correction \cite{yi2019probabilistic} 
                  & \small Best & \small 92.4 & \small 89.1 & \small 77.5 & \small 58.9 & \small 88.5\\
                  & \small Last & \small 92.0 & \small 88.7 & \small 76.5 & \small 58.2 & \small 88.3\\ \hline
    \small Meta-Learning \cite{li2019learning} 
                  & \small Best & \small 92.9 & \small 89.3 & \small 77.4 & \small 58.7 & \small 89.2\\
                  & \small Last & \small 92.0 & \small 88.8 & \small 76.1 & \small 58.3 & \small 88.6\\ \hline
    \small M-correction \cite{arazo2019unsupervised} 
                  & \small Best & \small 94.0 & \small 92.0 & \small 86.8 & \small 69.1 & \small 87.4\\
                  & \small Last & \small 93.8 & \small 91.9 & \small 86.8 & \small 68.7 & \small 86.3\\ \hline
    \small DivideMix \cite{li2020dividemix} 
                  & \small Best & \small 96.1 & \small 94.6 & \small 93.2 & \small 76.0 & \small 93.4\\
                  & \small Last & \small 95.7 & \small 94.4 & \small 92.9 & \small 75.4 & \small 92.2\\ \hline
    \small Ours          & \small Best & \small 90.1 & \small 81.8 & \small 64.4 & \small 46.2 & \small 88.6\\
    (default)     & \small Last & \small 88.8 & \small 81.6 & \small 56.1 & \small 39.8 & \small 88.4\\ \hline
    \small Ours          & \small Best & \small 93.3 & \small 91.1 & \small 77.1 & \small 52.2 & \small 91.2\\
    \small w/o noisy branch & \small Last & \small 93.1 & \small 90.8 & \small 76.5 & \small 45.6 & \small 90.9\\ \hline
    \footnotesize Ours w/o noisy branch  & \small Best & \small 93.9 & \small 91.8 & \small 81.3 & \small 61.6 & \small 92.1\\
    \footnotesize + Mixup & \small Last & \small 93.4 & \small 91.2 & \small 80.4 & \small 59.8 & \small 91.8\\ \hline
    \footnotesize Ours w/o noisy branch 
                  & \small Best & \small 94.2 & \small 93.9 & \small 92.5 & \small 76.2 & \small 92.3\\
    \footnotesize + Mixup + Soft Label & \small Last & \small 94.0 & \small 93.7 & \small 92.3 & \small 73.6 & \small 92.1\\ \hline
    \end{tabular}
    \label{fig:s3}
\end{table}

\begin{table}[t]
    \centering
    \caption{Test Accuracy (\%) of experiments on Cifar-100 with symmetric noise. Note that 20, 50, 80, and 90 denote noise rates; and, values in \textbf{Last} rows are the mean test accuracy values derived by models of last 10 training epochs.}
    \begin{tabular}{|l|l|r|r|r|r|}
    \hline
    \multicolumn{2}{|r|}{Noise Level (\%)}
                  & 20 & 50 & 80 & 90 \\ \hline
    Cross Entropy & \small Best & 62.0 & 46.7 & 19.9 & 10.1\\
                  & \small Last & 61.8 & 37.3 & 8.8 & 3.5 \\ \hline
    Mixup \cite{zhang2017mixup} & Best & 67.8 & 57.3 & 30.8 & 14.6 \\
                  & \small Last & 66.0 & 46.6 & 17.6 & 8.1 \\ \hline 
    P-correction \cite{yi2019probabilistic} 
                  & \small Best & 69.4 & 57.5 & 31.1 & 15.3 \\
                  & \small Last & 68.1 & 56.4 & 20.7 & 8.8 \\ \hline
    Meta-Learning \cite{li2019learning} 
                  & \small Best & 68.5 & 59.2 & 42.4 & 19.5 \\
                  & \small Last & 67.7 & 58.0 & 40.1 & 14.3 \\ \hline
    M-correction \cite{arazo2019unsupervised} 
                  & \small Best & 73.9 & 66.1 & 48.2 & 24.3 \\
                  & \small Last & 73.4 & 65.4 & 47.6 & 20.5 \\ \hline
    DivideMix \cite{li2020dividemix} 
                  & \small Best & 77.3 & 74.6 & 60.2 & 31.5 \\
                  & \small Last & 76.9 & 74.2 & 59.6 & 31.0 \\ \hline
    Ours          & \small Best & 65.6 & 50.3 & 23.1 & 10.4 \\
    (default)     & \small Last & 65.0 & 49.8 & 21.6 & 8.2 \\ \hline
    Ours          & \small Best & 74.1 & 65.9 & 40.8 & 15.2 \\
    w/o noisy branch & \small Last & 73.8 & 65.5 & 40.3 & 13.2 \\ \hline
    \small Ours w/o noisy branch  & \small Best & 76.1 & 70.2 & 45.3 & 21.8 \\
    \small + Mixup & \small Last & 75.9 & 69.8 & 44.6 & 19.9 \\ \hline
    \small Ours w/o noisy branch 
                  & \small Best & 76.6 & 73.9 & 57.2 & 28.6 \\
    \small + Mixup + Soft Label & \small Last & 76.4 & 73.1 & 56.9 & 28.4 \\ \hline
    \end{tabular}
    \label{fig:s5}
\end{table}

%% file: TMM_sec05_conclusion.tex
We proposed  an ensemble learning method for noisy label correction by using local patches of feature manifold. Based on the facts that most real-world noisy labels come from confusing label noises, we have proposed a nearest-neighbor-based data splitting scheme to tackle noisy labels locally-concentrated near decision boundaries by capturing the local structures of data manifolds . 
Hence, each sub-model can learn a coarse representation of the feature manifold, and only a few sub-models will be affected by locally-dense noisy labels. We have also proposed a multi-graph label propagation algorithm to derive reasonable label correction suggestions, and a confidence-guided label correction mechanism to determine proper label from the correction suggestions via majority decision. 
Extensive experiments and in-depth analyses on real-world datasets evidently demonstrate the superiority of the proposed method over existing noisy label correction strategies.



